\documentclass[conference]{IEEEtran}
\IEEEoverridecommandlockouts
\usepackage{wrapfig}
\usepackage{cite}
\usepackage{amsmath,amssymb,amsfonts}
\usepackage{algorithmic}
\usepackage{graphicx}
\usepackage{textcomp}
\usepackage{xcolor}
\usepackage{booktabs}
\usepackage{graphicx}
\usepackage{multirow}
\usepackage{url}
\usepackage{booktabs}
\usepackage{amssymb}
\usepackage{pifont}
\def\BibTeX{{\rm B\kern-.05em{\sc i\kern-.025em b}\kern-.08em
    T\kern-.1667em\lower.7ex\hbox{E}\kern-.125emX}}
\begin{document}

\title{SPT: Sequence Prompt Transformer for Interactive Image Segmentation\\
}

\author{
\centerline{
Senlin Cheng\textsuperscript{\rm 1}\textsuperscript{\#} \quad
Haopeng Sun\textsuperscript{\rm 3,4}\textsuperscript{\#} \quad
Tao Xie\textsuperscript{\rm 2}\textsuperscript{*}\quad
Hangyue Zhao\textsuperscript{\rm } \quad
Yiqiang Chen\textsuperscript{\rm 3,4} \textsuperscript{\Letter} \quad
Bolei Xu\textsuperscript{\rm 1} \quad
XiaoBo Li\textsuperscript{\rm 1} \quad
}\\

\centerline{
\textsuperscript{\rm 1} Ant Group, Hangzhou, China
\quad
\textsuperscript{\rm 2} BAIDU, Beijing, China} \\
\centerline{
\textsuperscript{\rm 3} Beiiing Key Laboratory of Mobile Computing and Pervasive Device, }\\
\centerline{
Institute of Computing Technology, Chinese Academy of Sciences, Beijing, China}\\
\centerline{
\textsuperscript{\rm 4}  University of Chinese Academy of Sciences, Beijing, China}\\

\centerline{
\url{senlin.csl@antgroup.com} \quad 
\url{sunhaopeng22s@ict.ac.cn} \quad 
\url{xietao07@baidu.com} \quad
\url{zhaohangyuefr@126.com} }\\
\centerline{
\url{yqchen@ict.ac.cn} \quad
\url{shishi.yl@antgroup.com} \quad
\url{xiaobo.lixb@antgroup.com} \quad
}}

\author{\IEEEauthorblockN{1\textsuperscript{st} Senlin Cheng}
\IEEEauthorblockA{\textit{Ant Group} \\
Hangzhou, China \\
senlin.csl@antgroup.com}
\and
\IEEEauthorblockN{2\textsuperscript{nd} Haopeng Sun}
\IEEEauthorblockA{\textit{Institute of Computing Technology, Chinese Academy of Sciences } \\
Beijing, China \\
sunhaopeng22s@ict.ac.cn}}







\maketitle

\begin{abstract}
Interactive segmentation aims to extract objects of interest from an image based on user-provided clicks. In real-world applications, there is often a need to segment a series of images featuring the same target object. However, existing methods typically process one image at a time, failing to consider the sequential nature of the images. To overcome this limitation, we propose a novel method called Sequence Prompt Transformer (SPT), the first to utilize sequential image information for interactive segmentation. Our model comprises two key components: (1) Sequence Prompt Transformer (SPT) for acquiring information from sequence of images, clicks and masks to improve accurate. (2) Top-k Prompt Selection (TPS) selects precise prompts for SPT to further enhance the segmentation effect. Additionally, we create the ADE20K-Seq benchmark to better evaluate model performance.
We evaluate our approach on multiple benchmark datasets and show that our model surpasses state-of-the-art methods across all datasets.
\end{abstract}

\begin{IEEEkeywords}
computer vision, interactive image segmentation
\end{IEEEkeywords}

\section{Introduction}
Interactive image segmentation leverages clicks between the user and the system to precisely segment target objects. Since segmentation can be achieved with just a simple click, it plays an increasingly important role in pixel-level annotation as well as in selecting objects of interest for editing and more~\cite{huang2023interformer,liu2023simpleclick}. In real-world applications, we often encounter the segmentation of parts of objects in a series of images, such as car windows, cat ears, human hands, etc. When segmenting car windows, the result might depict the whole car. To enhance precision, further refinement is necessary, such as adjusting boundaries or adding clicks. Existing methods~\cite{sofiiuk2022reviving,chen2022focalclick,liu2022pseudoclick} process one image at a time and do not utilize information from previous images in the sequence. When dealing with a series of images, existing methods that require corrections for each image are time-consuming and resource-intensive. Additionally, the timing and scenes in different images are inconsistent, making video-based interactive segmentation methods~\cite{cheng2021modular,miao2020memory} inapplicable.
\begin{figure}[t]
  \centering
    \includegraphics[width=0.7\linewidth]{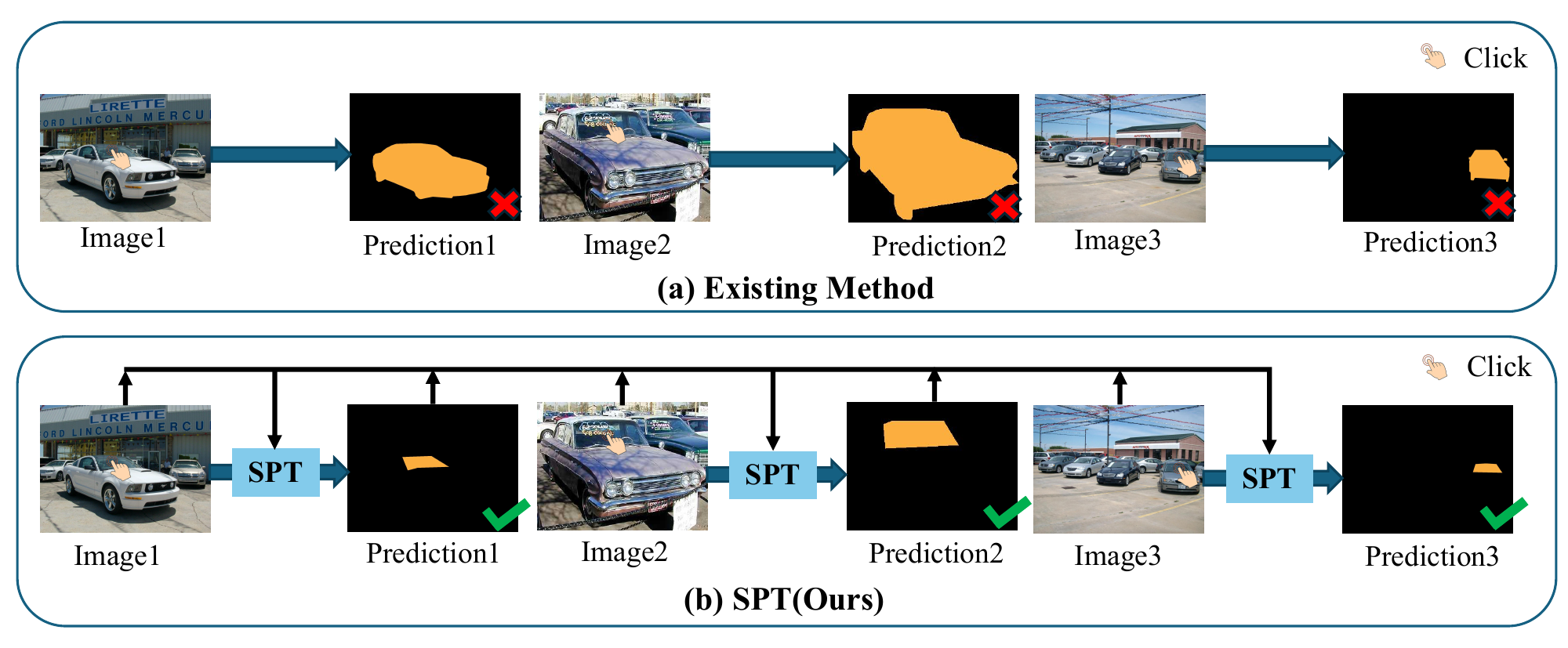}
   \caption{Segmentation of car windows: (a) Existing methods process individual images, causing prediction faults; (b) SPT learns useful information from previous images, clicks, and masks to achieve precise results. }
   \label{fig1}
\end{figure}

\begin{figure*}[tb]
    \begin{minipage}{0.55\textwidth}
        \centering
        \includegraphics[width=1.0\linewidth]{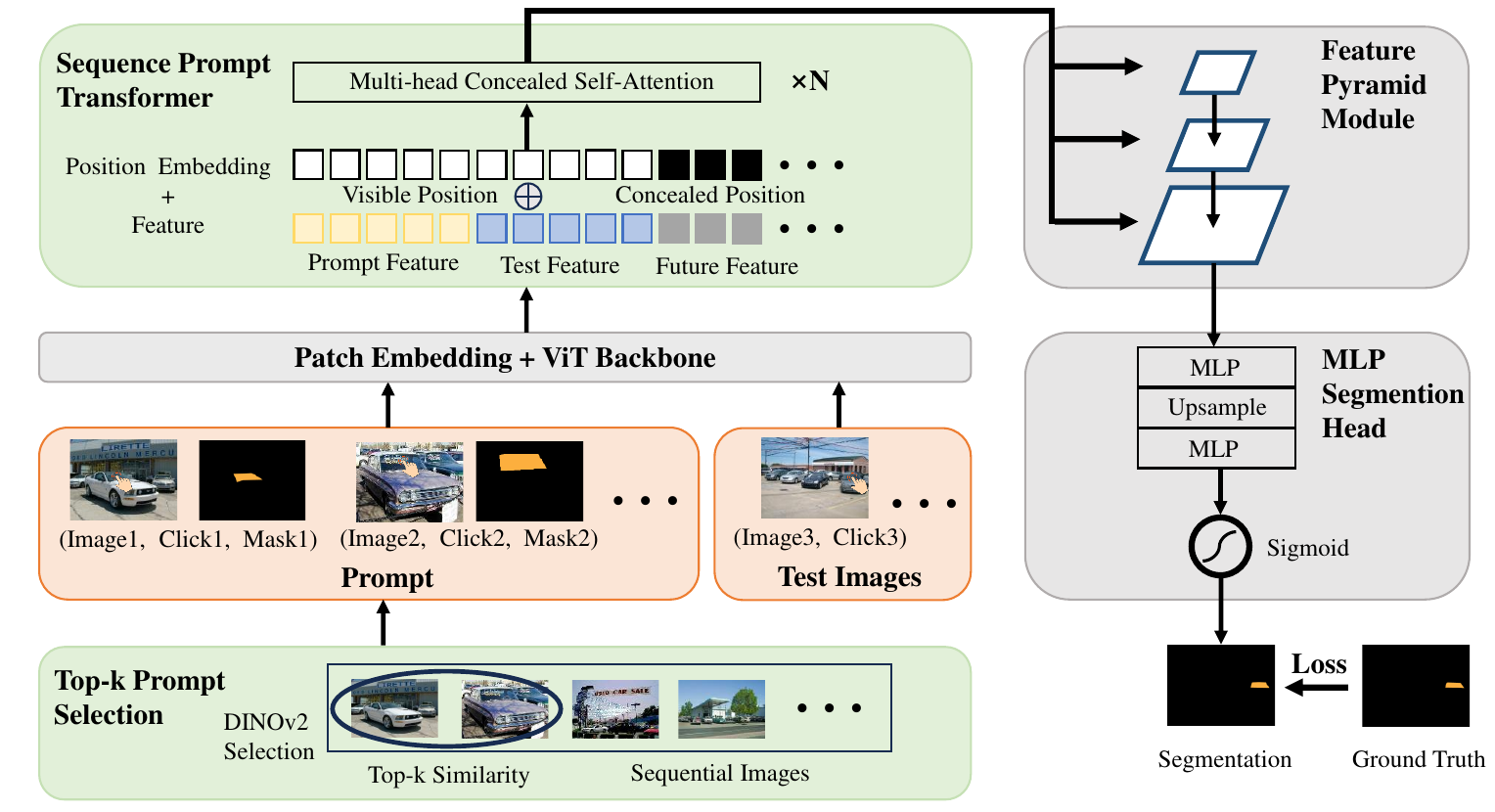}
        \caption{Overview of Sequence Prompt Transformer (SPT).
        }
        \label{fig2}
    \end{minipage}
    \hfill
    \begin{minipage}{0.43\textwidth}
        \centering
        \includegraphics[width=1.0\linewidth]{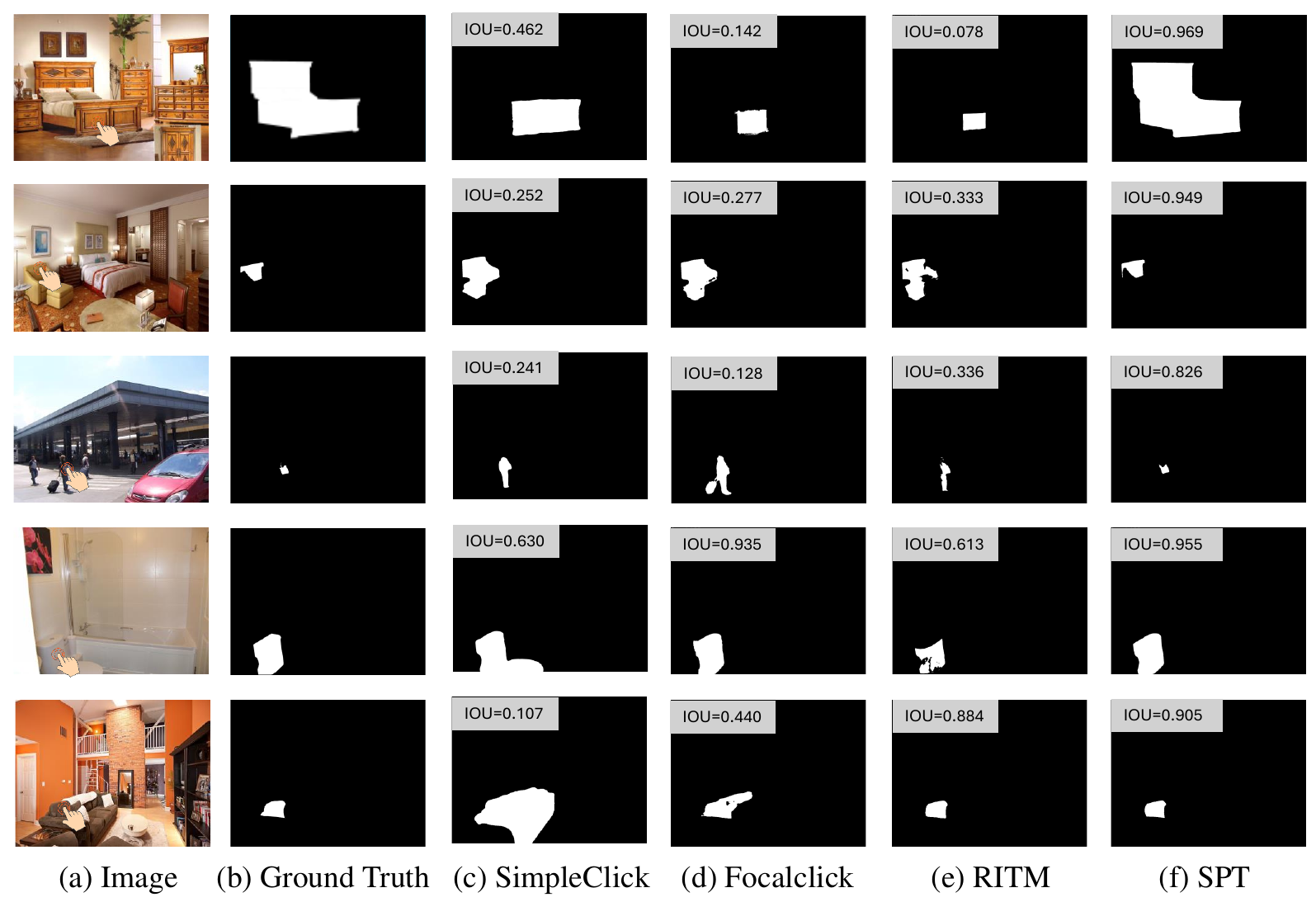}
        \caption{Qualitative analysis on the ADE20K-Sep dataset. 
    (a) Image. (b) Ground-truth mask. (c) Results of SimpleClick. (d) Results of Focalclick. (e) Results of RITM. (e) Results of SPT (ours).}
        \label{fig3}
    \end{minipage}
\end{figure*}


The human visual system first focuses on objects that are the same as those in previous images. Typically, sequence images of the same category, along with their clicks and predicted masks, contain similar information. Inspired by this, we propose the Sequence Prompt Transformer (SPT), which uses them as prompts to achieve more accurate segmentation results. As shown in Fig.~\ref{fig1}, when we attempt to segment a car window, existing methods might segment the entire car instead. 
Our method processes sequential images containing objects of the same category and learns useful information from previous images, clicks, and masks. Compared to existing methods, ours can accurately predictict car window, delivering better segmentation results. Different images can vary significantly, and generally, images with closer features are more suitable as prompts. To better learn the previous sequence information, we introduce the Top-k Prompt Selection (TPS), utilizing feature extraction as the prompt selection method and selects the top-k most similar images as prompts. We use DINOv2~\cite{oquab2023dinov2} to extract features from the sequence images and select the most similar images, clicks and masks as prompts. Extensive experiments demonstrate that our proposed TPS module significantly improves prediction accuracy, highlighting its effectiveness in enhancing prompt selection.

However, existing public datasets do not cater well to our proposed methodology. Predominantly, they are tailored for single-image scenarios, lacking a focus on sequential images. To bridge this gap, we introduce the ADE20K-Seq dataset, derived from the ADE20K~\cite{zhou2017ade20k} dataset, specifically designed to align with our task. We annotate the data according to the task level, We randomly select multiple tasks for testing, with ADE20K-Seq containing seven random tasks.

We conduct extensive experiments to evaluate the performance of our proposed Sequence Prompt Transformer (SPT) approach on five interactive segmentation datasets (i.e., GrabCut~\cite{rother2004grabcut}, Berkeley~\cite{mcguinness2010berkeley}, COCO-MVal~\cite{lin2014coco}, DAVIS~\cite{perazzi2016DAVIS} and ADE20K-Seq). Extensive experiments show that our proposed SPT outperforms other state-of-the-art methods on these datasets. 

Our main contributions are as follows: 
\begin{itemize} 
\item We propose a novel Sequence Prompt Transformer (SPT) for interactive image segmentation, leveraging sequential images, clicks, and predicted masks information as prompts to enhance segmentation accuracy.
\item In order to select the most appropriate prompt, we introduce the Top-k Prompt Selection (TPS) module. TPS uses DINOv2 for feature extraction, selecting the top-k most similar images, clicks, and masks as prompts for more accurate segmentation.
\item To better evaluate our task, we introduce the ADE20K-Seq dataset, the first dataset dedicated to interactive segmentation of sequential images from the same category. This dataset is rich in data, including seven different categories at various scales, each with over 100 images.
\item Extensive experiments conducted on popular interactive segmentation benchmarks demonstrate the superiority of SPT over previous state-of-the-art methods. The results validate the effectiveness of SPT and TPS in addressing interactive image segmentation tasks.

\end{itemize}

\section{Related work}
Interactive image segmentation is a longstanding problem for which increasingly better solution approaches have been proposed. Early research~\cite{boykov2001interactive,gulshan2010geodesic,rother2004grabcut} tackled this issue using graphs defined over image pixels. However, these methods only focused on low-level image features, often struggling with complex objects. With the advent of deep learning~\cite{vaswani2017attention} and large datasets, modern approaches have greatly improved the quality of interactive segmentation. CDNet~\cite{chen2021cdnet} employs self-attention to better focus on relevant image regions. RITM~\cite{sofiiuk2022reviving} improves segmentation with a click-based interaction strategy. FocalClick~\cite{chen2022focalclick} developes iterative methods to enhance mask quality, and Pseudoclick~\cite{liu2022pseudoclick} reduces user input by predicting future clicks. SimpleClick~\cite{liu2023simpleclick} utilizes the Vision Transformer (ViT)~\cite{dosovitskiy2020image} backbone to leverage pre-trained weights. The Segment Anything Model (SAM)~\cite{kirillov2023segment} is a zero-shot model that uses a large dataset to handle diverse objects and scenes. HQ-SAM~\cite{ke2023hq-sam} builds on this with high-quality output tokens and global-local feature fusion, improving accuracy for complex objects. These models have significantly enhanced the efficiency, accuracy, and applicability of interactive segmentation. However, existing methods ignore the sequential information in sequential images, leading to errors in object segmentation. To our knowledge, our method is the first to incorporate sequential information into interactive image segmentation, effectively improving accuracy.

\section{Method}
The overall structure of our approach is demonstrated in Fig.~\ref{fig2}. Our model input consists of a sequence of images from the same category. We use Top-k Prompt Selection (TPS) to choose the most suitable images as prompts. We concatenate and encode the 'click' and 'mask' information along with the images. The encoded information is passed through the Patch-Embedding and ViT network for feature extraction and subsequently fed into the Sequence Prompt Transformer (SPT) to capture inter-image relationships within the sequence. The features are then processed through the Feature Pyramid Module (FPM) and MLP Segmentation Head (MSH) to produce the predicted segmentation result.

\subsection{Network Input}
We aim to tackle the challenge of segmenting a sequence of images that require labeling the same type of object. We select an instance from the dataset and search for multiple images belonging to the same category. The network input consists of a sequence of images with corresponding clicks and masks, denoted as $S_{input}$, and can be expressed as:
\[
S_{input} = \{(I_1, C_1, M_1), (I_2, C_2, M_2), \dots, (I_i, C_i, M_i)\},
\]
where $I_i$ represents the images, $C_i$ represents the clicks on image $I_i$, and $M_i$ is the mask generated for $I_i$ by the network. To closely simulate the real experience of users making iterative clicks on images, we utilize the click simulation strategy described in RITM. This process continues for each image until either the number of clicks reaches a predefined maximum, or the Intersection over Union (IoU) surpasses a certain threshold, indicating satisfactory alignment with the ground truth.
\subsection{Backbone}
We use ViT~\cite{dosovitskiy2020image} as the backbone network. For each pair in $S_{input}$, we concatenate $I_i$, $C_i$, and $M_i$ and pass them through a patch embedding layer. The resulting embedded representation is element-wise added to the embedded representation of the image. This summed representation is then fed into the ViT to obtain the feature $F_i$. Mathematically, this can be expressed as:
$$F_i=ViT(Embed(C_i \oplus M_i)+Embed(I_i)),$$where $\oplus$ denotes the concatenation operation, and $Embed$ represents the patch embedding function. After processing through the backbone, the feature sequence $F$ is obtained:
 $$F=\{F_1,F_2,F_3, \cdots,F_i\}.$$ 
\subsection{Top-k Prompt Selection (TPS)}
The sequence of images, along with the click and mask information, is used to guide the prediction and segmentation of subsequent images, allowing the network to continuously learn contextual information throughout the sequence. However, due to the large number of images in the sequence, using all of them as prompt inputs is impractical. To address this, we propose a Top-k Prompt Selection (TPS) strategy to choose the most appropriate $k$ prompts. Typically, images that are more similar to the test image are better suited as prompts. We use DINOv2 model to measure the similarity between different images and select the $k$ most similar ones as prompt inputs.

\subsection{Sequence Prompt Transformer (SPT)}
We use sequence images, clicks, and predicted masks as prompts to guide the segmentation of test images. To enable the network to integrate attributes from prior inputs and previous predictions within the sequence, we introduce the Sequence Prompt Transformer (SPT) after the backbone. In processing sequences within the Transformer architecture, it is essential to ensure that information from future positions remains inaccessible at the current position. To achieve this, we propose a Multi-head Concealed Self-Attention mechanism, where images beyond the current test image in the sequence are concealed. We introduce position coding to indicate whether a token is visible, ensuring that each token can only attend to preceding positions and itself, thereby preventing the learning of future information. Given the feature sequence $F$, it is processed through the SPT to yield the feature sequence $F_s$:

\[
F_s[i] = \text{SPT}(F_i \oplus \text{mask}(i)),
\]

where the mask function for an element at position $i = (x, y)$ in the sequence is defined as:

\[
\text{mask}(x, y) = \left\{
\begin{aligned}
1 & \quad \text{if } (x \ge y) \\
0 & \quad \text{otherwise}.
\end{aligned}
\right.
\]
This means that only the current position and earlier positions are visible, while positions after the current one remain invisible. The SPT comprises position embedding and $N$ Multi-head Concealed Self-Attention layers.

\subsection{Loss Function}

The feature sequence \( F_s \) is passed through the Feature Pyramid Module (FPM) and the MLP Segmentation Head (MSH) to obtain the predicted result \( S_{pred} \). FPM can integrate features of different scales to help capture the fine details of small objects in the image while maintaining a global perception of large objects. MSH is a lightweight segmentation head that uses only MLP layers to avoid computationally demanding components, accounting for only up to 1\% of the model parameters. The loss is then computed by comparing \( S_{pred} \) with the ground truth \( S_{GT} \). This process is represented as:

\[
S_{pred} = \text{MSH}(\text{FPM}(F_s)).
\]

The loss function \( L \) is given by:

\[
L = \text{FocalLoss}(S_{pred}, S_{GT}).
\]

where the Focal Loss~\cite{lin2017focal} is defined as:

\[
\text{FocalLoss}(p_t) = -\alpha (1 - p_t)^\gamma \log(p_t),
\]

with

\[
p_t = \left\{
\begin{aligned}
S_{pred} & \quad \text{if foreground} \\
1 - S_{pred} & \quad \text{otherwise}.
\end{aligned}
\right.
\]

\section{Experiment}
\subsection{Experimental Setup}
\subsubsection{Dataset}
We train our model on the COCO~\cite{lin2014coco} (118k training images and 1.2M instances) and LIVIS~\cite{gupta2019lvis} (100k images and 1.2M instances) datasets. We evaluate our approach on four benchmarks: GrabCut~\cite{rother2004grabcut} contains 50 images, each with only one instance; Berkeley~\cite{mcguinness2010berkeley} contains 96 images with 100 instances; COCO-MVal~\cite{lin2014coco} extends the COCO dataset by providing click-based annotations for objects, offering a diverse range of scenes and object categories for interactive segmentation research; and the DAVIS dataset~\cite{perazzi2016DAVIS}, known for its high-quality video annotations, emphasizes temporally consistent segmentations.

\subsubsection{New Benchmark: ADE20K-Seq}
Existing public datasets for interactive segmentation primarily focus on evaluations based on individual images. To address the lack of sequential segmentation datasets, we propose ADE20K-Seq dataset. We extend ADE20K dataset into 7 category-specific benchmarks, with each category containing more than 100 images, thereby constructing the ADE20K-Seq dataset. The ADE20K-Seq dataset is notable for its broad range of categories, including large-scale objects such as airplanes and beds, as well as smaller items like pillows, mugs, and balls.

\subsubsection{Implementation}
We rescale images by 0.7 to 1.4 and use data augmentation techniques like random flipping and brightness enhancement. Two ViT models, ViT-B and ViT-H, initialized with MAE pre-trained weights, serve as the backbone. We train the model for 55 epochs on COCO and LVIS datasets using the Adam optimizer ($\beta_1 = 0.9$, $\beta_2 = 0.999$), with an initial learning rate of $5 \times 10^{-4}$, reduced to $5 \times 10^{-6}$ after 50 epochs. Training is performed on four A100 GPUs, with batch sizes of 16 for ViT-B and 8 for ViT-H. The prompt sequence length is 10.

\subsubsection{Evaluation Metric}
Following prior works~\cite{sofiiuk2022reviving,liu2023simpleclick,chen2022focalclick}, we use "NoC" (Number of Clicks) as our evaluation metric. NoC85 indicates the number of clicks required to achieve an 85\% Intersection over Union (IoU) score, while NoC90 is for a 90\% IoU. These metrics provide a clear measure of the efficiency and accuracy of the methods.

\subsection{Experimental Results}
We present a comprehensive evaluation of our SPT framework in comparison to state-of-the-art models across five benchmark datasets. Table~\ref{sota_compare} highlights the overall performance of our method against baselines including RITM, EdgeFlow, CDNet, PseudoClick, Focalclick, EMC, SAM, InterFormer, SimpleClick, and HQ-SAM. Our SPT framework consistently outperforms the baselines across all datasets and evaluation metrics, achieving state-of-the-art results. These findings underscore the effectiveness and robustness of our approach in addressing the challenges posed by interactive image segmentation.

\begin{figure}[t]
  \centering
    \includegraphics[width=0.65\linewidth]{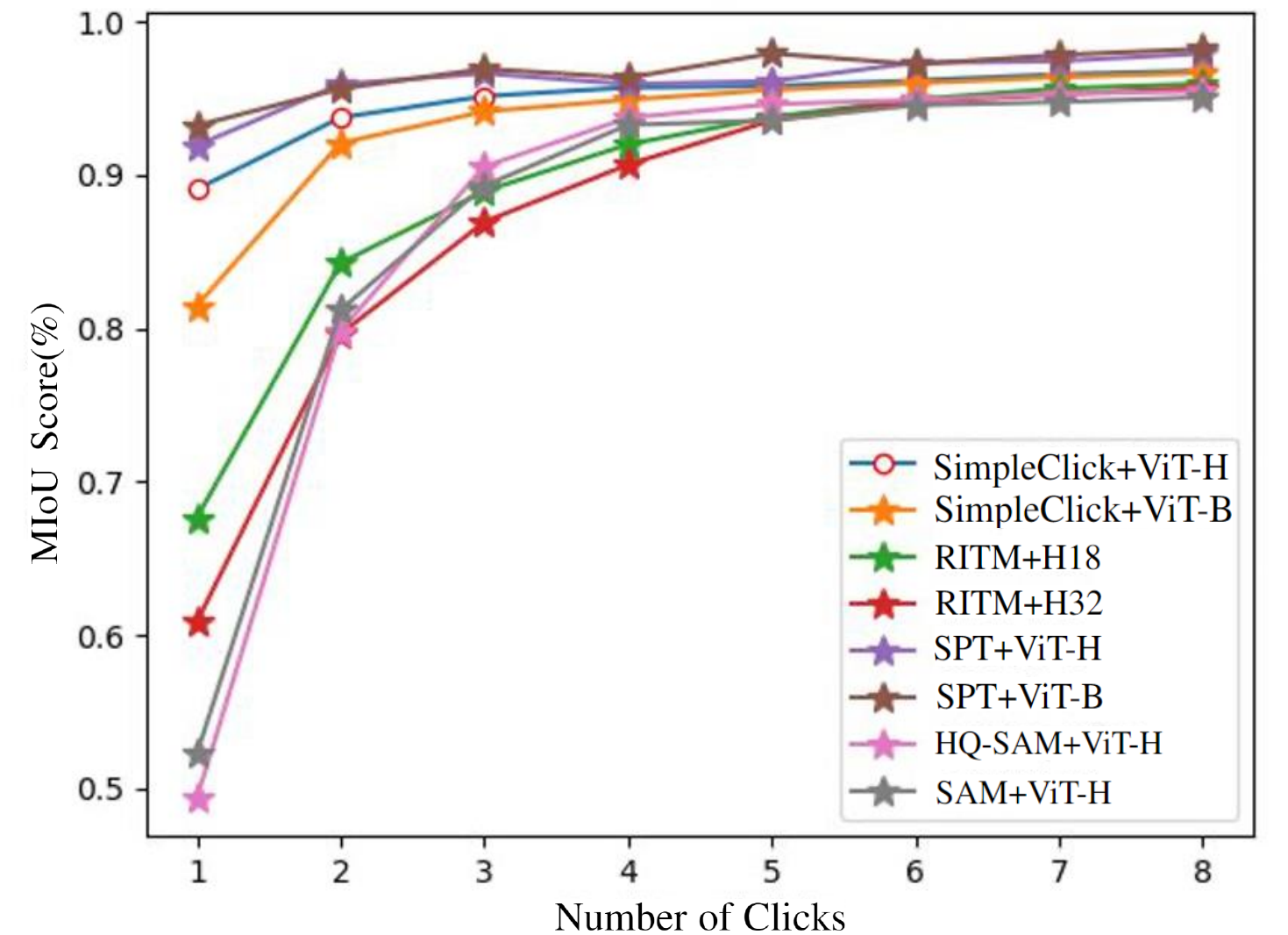}
   \caption{Comparison of MIoU performance with different numbers of clicks against baselines on the ADE20K-Sep dataset. MIoU higher is better.}
   \label{fig4}
\end{figure}

Additionally, we conduct experiments with varying numbers of click inputs on ADE20K-Sep dataset. As shown in Fig.~\ref{fig4}, our method consistently achieves the best results, regardless of the number of input clicks (represented by the purple and brown lines). Notably, with fewer clicks, our method significantly outperforms the baseline.

\subsection{Ablation Study}

We present ablation studies in Table~\ref{ablation} to evaluate the effectiveness of the proposed Sequence Prompt Transformer (SPT) and Top-k Prompt Selection (TPS) on the GrabCut, Berkeley, COCO MVal, DAVIS datasets.

\textbf{Effect of Sequence Prompt Transformer (SPT).} 
To demonstrate the effectiveness of our proposed SPT method, we compare it with prompt sequences of different lengths. As shown in ExpID \#1 (baseline without prompt) and \#5 (our method SPT setting), the proposed SPT effectively improves the performance of interactive segmentation. We also test different prompt lengths, as shown in ExpID \#2, \#3, \#4, and \#5. As the length of the prompts increases, the model's performance continuously improves, further demonstrating the benefits of prompts for interactive segmentation. Under our setting, the best performance is achieved (Exp \#5).

\textbf{Effect of Top-k Prompt Selection (TPS).} As shown in ExpID \#6 (without prompt selection) and \#9 (our TPS method), TPS significantly enhances model performance, highlighting the importance of prompt selection. We also test different selection models and demonstrate that our proposed method is the most effective, leading to improved accuracy (ExpID \#6, \#7, \#8 and \#9).

\textbf{Qualitative Analysis.}
To demonstrate the effectiveness of our proposed method, we conduct a qualitative analysis on the ADE20K-Seq dataset in Fig.~\ref{fig3}.
We visualize the source image, the ground-truth mask, the results of Simpleclick, Focalclick, RITM and our proposed SPT.
We compare our SPT with the previous state-of-the-art methods. Our SPT produces more precise results with better semantics and finer segmentation boundaries, showcasing its superior performance.
\section{Conclusion}
In this work, we propose a novel interactive image segmentation method, termed the Sequence Prompt Transformer (SPT). The Sequence Prompt Transformer effectively captures information from sequences of images, clicks, and masks. The Top-k Prompt Selection (TPS) enhances segmentation results by selecting more suitable prompts. Additionally, we introduce a new dataset for sequence-based interactive image segmentation. Extensive experiments demonstrate that our SPT consistently outperforms existing methods on various interactive image segmentation benchmarks.


\bibliographystyle{IEEEtran}

\end{document}